\documentclass{article} 
\usepackage[preprint]{colm2025_conference}

\usepackage{microtype}
\usepackage{hyperref}
\usepackage{url}
\usepackage{booktabs}
\usepackage{graphicx}
\usepackage[para,online,flushleft]{threeparttable}
\usepackage{multirow}
\usepackage{lineno}
\usepackage{lettrine,Zallman}
\usepackage{algorithm}
\usepackage{algpseudocode}
\usepackage{adjustbox}
\usepackage{diagbox}

\definecolor{darkblue}{rgb}{0, 0, 0.5}
\hypersetup{colorlinks=true, citecolor=darkblue, linkcolor=darkblue, urlcolor=darkblue}

\title{FRASE: Structured Representations for Generalizable SPARQL Query Generation}

\author{Papa Abdou Karim Karou Diallo$^{\diamond \dagger \star}$ ~~~ Amal Zouaq$^{\diamond \dagger \star}$\\ \\
\medskip
$^{\diamond}$LAMA-WeST  ~~~~~~ $^{\dagger}$Polytechnique Montreal ~~~~~~ $^{\star}$Mila\\
\texttt{\{diallokarou28, amal.zouaq\}@polymtl.ca} \\
}

\begin{document}
\ifcolmsubmission
\linenumbers
\fi

\maketitle

\begin{abstract}
Translating natural language questions into SPARQL queries enables Knowledge Base querying for factual and up-to-date responses. However, existing datasets for this task are predominantly template-based, leading models to learn superficial mappings between question and query templates rather than developing true generalization capabilities. As a result, models struggle when encountering naturally phrased, template-free questions. This paper introduces FRASE (\underline{FRA}me-based \underline{S}emantic \underline{E}nhancement), a novel approach that leverages Frame Semantic Role Labeling (FSRL) to address this limitation. We also present LC-QuAD 3.0, a new dataset derived from LC-QuAD 2.0, in which each question is enriched using FRASE through frame detection and the mapping of frame-elements to their argument. We evaluate the impact of this approach through extensive experiments on recent large language models (LLMs) under different fine-tuning configurations. Our results demonstrate that integrating frame-based structured representations consistently improves SPARQL generation performance, particularly in challenging generalization scenarios when test questions feature unseen templates (unknown template splits) and when they are all naturally phrased (reformulated questions).
\end{abstract}

\vspace{5ex}
\section{Introduction}
\label{sec:introduction}

\lettrine[lines=2]{I}{}nformation democratization aims to ease access to the vast amount of factual information stored in Knowledge Bases (KBs) and relational databases. A crucial step towards this goal is the translation of natural language questions into structured queries such as SPARQL, SQL or S-expression \cite{diallo2024comprehensive, banerjee2022modern,sharma2025reducing}. 
Recent advances in Large Language Models (LLMs), particularly decoder-only architectures, have significantly improved the semantic representation of natural language, achieving state-of-the-art results on a variety of natural language understanding tasks. 

Nevertheless, despite these successes, LLMs remain sensitive to input phrasing and prompt formulation, resulting in limited robustness and generalization across diverse question formulations and unseen patterns \citep{leidinger2023language, zheng2023prompt}.
A key limitation of current question-to-SPARQL systems lies in the datasets on which they are trained \cite{diallo2024comprehensive, reyd2023assessing}. Many benchmarks rely on rigid, template-based constructions, which lead models to learn surface-level mappings between input questions and query structures. This shortcut learning hinders their ability to generalize to naturally phrased, template-free questions, especially when the input deviates from patterns seen during training.

To address these aforementioned challenges, we introduce \textbf{FRASE} (\underline{FRA}me-based \underline{S}emantic \underline{E}nhancement), a method that augments natural language questions with structured semantic information derived from Frame Semantics. These Frames have proven valuable in enhancing semantic understanding in tasks such as machine reading comprehension \citep{guo2020incorporating, flanigan2022meaning, bonn2024meaning} and information extraction \citep{su2021knowledge, li2024comprehensive, chanin2023open, su2023span}.

\begin{figure}
\centering
\includegraphics[width=0.8\textwidth]{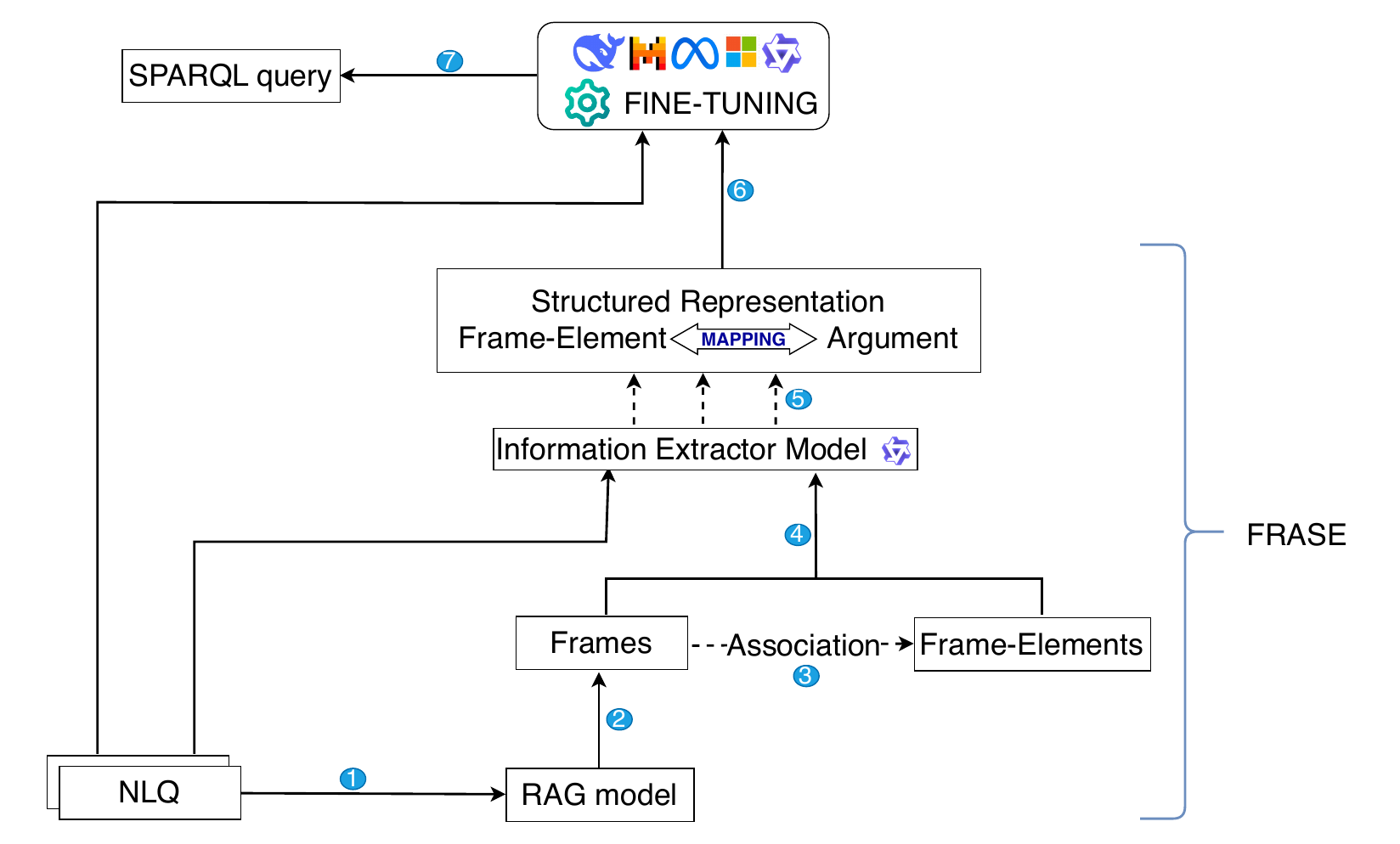}
\caption{Overview of our proposed architecture.}
\label{fig:pipeline}
\end{figure}

As illustrated in Figure \ref{fig:pipeline}, \textbf{FRASE} leverages a two-stage pipeline to detect the frames evoked by the question and identify their associated semantic roles, which are then used to enrich the question representation. A detailed presentation of the process is shown in Section \ref{sec:frase-main-architecture}. 
We hypothesize that integrating such structured representation into question-to-query models can improve their ability to abstract and generalize beyond rigid surface forms. To evaluate this hypothesis, we introduce LC-QuAD3.0, a new dataset derived from LC-QuAD2.0 \cite{LCQUAD2}, in which each question is annotated with its corresponding frame and frame-element and argument mapping using the \textbf{FRASE} pipeline. We conduct comprehensive experiments with several recent LLMs under different fine-tuning settings, assessing the impact of our semantic augmentation across various dataset splits, including out-of-distribution (OOD) settings.

Our contributions are threefold: 
\begin{enumerate} 
    \item We propose \textbf{FRASE}, a new method for frame detection and arguments identification that does not rely on manually identified target spans, leveraging a RAG-based system grounded in KB relation semantics and frame definitions. 
    \item We demonstrate that enriching questions with frame-based structured representations improves generalization in SPARQL query generation. 
    \item We show that this improvement holds not only for unseen-template test sets but also for challenging, naturally phrased reformulations. 
\end{enumerate}

\section{Related Works}
\label{sec:related-works}
\subsection{SPARQL query generation}
SPARQL generation from natural language has been extensively studied using both Small and Large Language Models (SLMs and LLMs) \citep{diallo2024comprehensive, reyd2023assessing, sharma2025reducing, banerjee2022modern, emonet2024llm, zahera2024generating}. These models are typically fine-tuned end-to-end, often with enhancements such as copy mechanisms \citep{banerjee2022modern, diallo2024comprehensive} or non-parametric memory modules \citep{sharma2025reducing} to reduce URI-related errors.

While many systems explicitly generate SPARQL queries, others bypass query generation altogether by having LLMs directly produce answers grounded in the knowledge base \citep{shavarani2024entity, alawwad2024enhancing, muennighoff2022sgpt}. Prompt engineering has also become a prominent strategy, using few-shot examples or explicit URI context to guide generation \citep{luo2023chatkbqa, muennighoff2022sgpt, diallo2024comprehensive}. Models such as Code Llama v2 \citep{roziere2023code}, Mistral 7B \citep{jiang2023mistral7b}, and Mistral 7B Instruct\footnote{\url{https://huggingface.co/mistralai/Mistral-7B-Instruct-v0.3}} have been widely adopted for such tasks.

However, despite promising results, existing methods often under-perform when faced with naturally phrased or paraphrased questions. Studies on LC-QuAD 2.0 show a marked drop in accuracy when models are evaluated on reformulated (template-free) questions, highlighting poor generalization to unseen linguistic patterns \citep{diallo2024comprehensive, reyd2023assessing}.

\subsection{Frame Semantic Parsing}

Frame semantic parsing approaches fall into two main categories: generative sequence-to-sequence methods and representation learning techniques leveraging FrameNet’s structured knowledge.

Seq2Seq models treat frame parsing as a generation task that includes frame identification and argument extraction \citep{sutskever2014sequence, raffel2020exploring, kalyanpur2020open, chanin2023open}. Architectures such as T5 \citep{raffel2020exploring}, pre-trained on PropBank \citep{kingsbury2002treebank} and FrameNet, are often used with lexical units and data augmentation to improve robustness. Some models adopt multi-task learning with shared encoders and task-specific decoders \citep{kalyanpur2020open}.

Representation learning methods, in contrast, focus on aligning sentence-level or span-level embeddings with candidate frames \citep{jiang2021exploiting}. These approaches often use Graph Neural Networks \citep{wu2020comprehensive} or contrastive learning \citep{ju2024towards, an2023coarse} to integrate semantic relations among frames, elements, and lexical triggers \citep{su2021knowledge, zheng2022double, tamburini2022combining}.

A major limitation of both paradigms is their reliance on an explicit target lexical unit—an annotation absent in datasets like LC-QuAD 2.0. Without this, frame-based parsing becomes inapplicable, and structured semantic representations cannot be extracted or evaluated in downstream tasks such as SPARQL generation.

\section{Methodology}\label{sec:methodology}
\subsection{Data}\label{sec:methodology-data}

\subsubsection{LC-QuAD2.0}\label{sec:methodology-data-lcq2}
The LC-QuAD 2.0 dataset \cite{LCQUAD2} (or LCQ2 for short) is composed of natural language questions paired with their corresponding SPARQL queries. Each entry includes two semantically equivalent versions of the question: one generated from a predefined template and another reformulated manually to resemble more natural human phrasing. This dual-question format enables nuanced evaluation of model generalization to both synthetic and naturally expressed inputs.
LCQ2 incorporates a wide range of SPARQL constructs, including advanced operators such as FILTER, which often require reasoning over literals—raw values in the knowledge graph such as strings, numbers, and dates. 

In our experiments, we consider two distinct data splits of LCQ2, following the protocol established by \citet{reyd2023assessing}:
\paragraph{Original Split.} This conventional split follows a standard 80-10-10 train/validation/test partitioning strategy, with no constraints on the overlap of question-query templates across sets. As a result, similar question-query structures may appear in both training and test sets, allowing models to benefit from direct pattern reuse.
\paragraph{Unknown Template Split.} To evaluate compositional generalization by ensuring that no templates seen in the test set appear in the training set, we group question-query pairs by their global templates and select test instances exclusively from groups containing templates that are absent from the training data. This separation results in fully disjoint template sets and imposes a more rigorous evaluation setting. Statistical analysis of these two splits is shown in Table \ref{tab:lcquad2-splits-stats}.

\begin{table*}
\centering
\begin{threeparttable}
\begin{adjustbox}{width=\linewidth}
\begin{tabular}{l l r r r r r}
\toprule
\multicolumn{2}{c}{} & \multicolumn{5}{c}{\textbf{Statistics}} \\
\cmidrule(lr){3-7}
\textbf{Dataset variant} & & \textbf{Total} & \textbf{Train} & \textbf{Validation} & \textbf{Test} & \textbf{Unseen} \\
\midrule
\multirow{3}{*}{Original Split} 
& Global templates    & 30    & 30    & 30    & 30    & 0    \\
& Entries             & 30225 & 21761 & 2418  & 6046  & 6046 \\
& Avg Query-Length  & - & 17 & 18  & 18  & - \\
\midrule
\multirow{3}{*}{Unknown Template Split} 
& Global templates    & 30    & 24    & 6    & 6    & 6    \\
& Entries             & 30225 & 24178 & 3023  & 3024  & 3024 \\
& Avg Query-Length  & - & 16 & 16  & 36  & - \\
\bottomrule
\end{tabular}
\end{adjustbox}
\begin{tablenotes}
     \item[] Query length is measured in term of number of words in it.
  \end{tablenotes}
\end{threeparttable}
\label{tab:lcquad2-splits-stats}
\caption{LC-QuAD 2.0 statistics in terms of global templates and entries across data splits.}
\end{table*}

To further analyze the influence of question formulation on model performance, we exploit the dual-question structure of LCQ2 to construct three dataset variants: \textbf{(Raw Questions}: Only the original, template-based questions are retained. \textbf{Reformulated Questions}: Only the human-written, naturally phrased questions are used. \textbf{Combined Questions}: Both question versions are treated as distinct entries, effectively doubling the dataset size and increasing question variety.
This experimental design allows us to address the following research questions:\\ 
\textbf{RQ1}: How does model performance vary when trained and evaluated on template-based questions versus naturally phrased or questions with unseen templates?\\
\textbf{RQ2}: Can combining template-based and template-free questions during training improve generalization and how effective is training exclusively on one of these two types ? \\
\textbf{RQ3}: To what extent does incorporating structured semantic representations based on frames improve performance across these different training and evaluation configurations?

\subsection{FRASE Main Architecture}\label{sec:frase-main-architecture}
The motivation behind incorporating structured semantic representations via Frame Semantic Role Labeling (FSRL) lies in the observation that different surface formulations of a question can often share the same underlying meaning. Regardless of phrasing, such questions typically evoke the same core event or concept, along with a consistent set of participants and their roles. This intuition aligns closely with the theory of frame semantics, where each frame represents a conceptual structure that encapsulates an event or situation, and frame elements denote the roles associated with its participants. 
Given an input text such as a natural language question, \textbf{FRASE} aims to extract its structured semantic representation by (1) identifying the frame(s) it evokes and (2) mapping the associated frame elements to their corresponding spans within the text. 

\subsubsection{Stage 1: Frame detection}\label{sec:frase-main-architecture-stage1}
In this stage, we employ a Retrieval-Augmented Generation (RAG) model to identify the frames evoked by each question in the LCQ2 dataset. Our core assumption is that, for any natural language question associated with a SPARQL query—hence linked to a set of knowledge base (KB) entities, relations and classes, we can identify a corresponding semantic frame that reflects the same conceptual structure as the KB relations/classes involved.
This assumption is grounded in the observation that both FrameNet frames and KB relations are associated with textual descriptions that encapsulate their underlying semantics. A well-trained embedding model is thus used to produce similar representations for these two descriptions, enabling us to establish a semantic alignment between the KB relation and the frame.
By leveraging this alignment through semantic similarity search, our RAG model retrieves the most relevant frames for a given question by comparing the question’s content to the descriptions of frames and KB relations/classes. This mechanism allows us to identify which frame is most likely evoked by the question, forming the first step in constructing a structured semantic representation.

\paragraph{Frame and KB Relation/Class Representations.} To enable effective semantic alignment between FrameNet frames and knowledge base (KB) relations/classes, we represent a frame in the most expressive and discriminative method as suggested by results in from \cite{diallo2025enhancing}. Such representation combines all three components: the frame \textit{label}, its \textit{definition} (or description), and the list of of its \textit{frame-elements}. This enriched representation captures both the semantic core of the frame and the structure of its participant roles, which improves retrieval performance. In contrast, representing KB relations/classes is more straightforward. For each relation/class URI in Wikidata, we concatenate the \textit{relation label} (i.e., its name) with its \textit{textual description} as provided by the KB. This representation succinctly captures the intended semantics of the relation/class and is well-suited for embedding-based similarity search.

\paragraph{Embedding Model and Semantic Retrieval.} We use the english version of BGE\footnote{https://huggingface.co/BAAI}\citep{li2024makingtextembeddersfewshot,bge_embedding} embedding-model to encode all frame representations into fixed-length vector embeddings, which are stored in a vector index to enable efficient retrieval. At inference time, for each question in LCQ2, we extract the textual descriptions of the relation/class URIs present in the corresponding SPARQL query. These relation/class descriptions are then embedded into the same vector space. Using these embeddings, we perform a top-\(k\) similarity search (with \(k = 1\) in the case of one-to-one alignment) to retrieve the most semantically similar FrameNet frame(s) from the vector store. 

\begin{figure}
\centering
\includegraphics[width=0.9\textwidth]{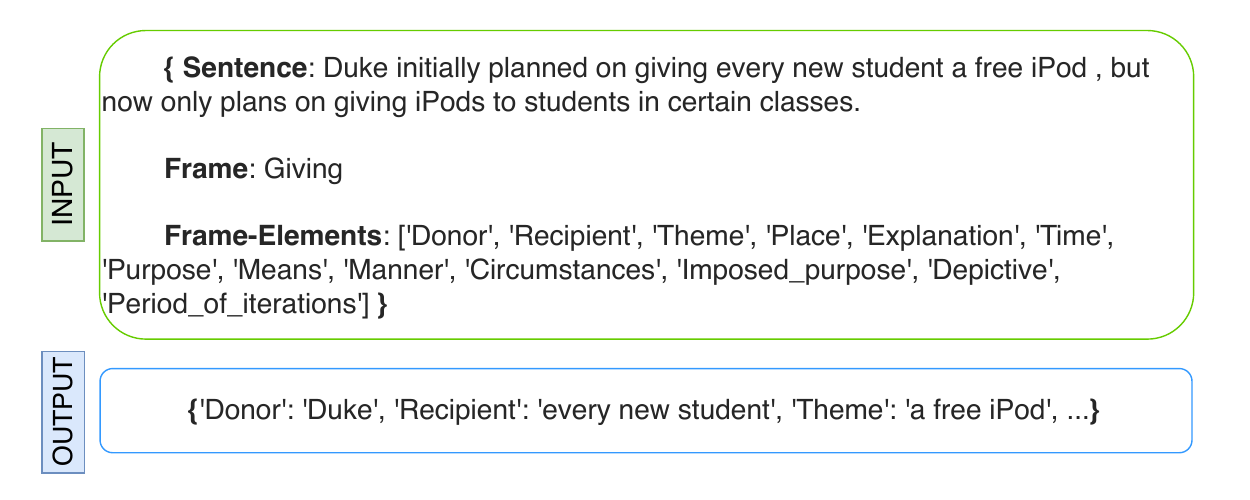}
\caption{An example of frame-based structured representation of text.}
\label{fig:annotated_example_FrameNet}
\end{figure}

\subsubsection{Stage 2: Arguments identification}\label{sec:frase-main-architecture-stage2}
Once the most semantically aligned frame has been retrieved for each KB relation/class in a given LCQ2 question, the next step involves identifying the arguments—that is, mapping each frame-element associated with the evoked frame to its corresponding span of text in the question. The frame alone provides the conceptual structure of the event or situation, but a complete semantic representation requires grounding the roles (frame elements) in the actual input question.
To accomplish this, we fine-tuned the model Qwen2.5-7B \cite{yang2024qwen2} which was empirically identified as one of the best-performing models for frame-semantic argument extraction in our preliminary experiments. The model is trained to generate structured outputs in a JSON format (see Figure \ref{fig:annotated_example_FrameNet}), following the recommendation and setup proposed by \citet{devasier2025llmsextractframesemanticarguments}. 
The fine-tuning process is carried out on the FrameNet dataset, using its fully annotated exemplars. The training split comprises \(3,353\) sentences, encompassing a total of \(19,391\) frame annotations and \(34,219\) frame elements. The test split includes \(1,247\) sentences, with \(6,714\) frames and \(11,302\) annotated frame elements. As a result of Stage 2, we obtain a model specifically trained to perform argument mapping. Given an input text and the set of frames evoked within it, identified during Stage 1 (see Section~\ref{sec:frase-main-architecture-stage1}), the model outputs a mapping between each frame-element and its corresponding text span in the input. The Figure \ref{fig:annotated_example_FrameNet} illustrates one example of this mapping from FrameNet data used for fine-tuning.

\subsubsection{Stage 3: LCQ3 Generation}
By applying \textbf{FRASE} to each LCQ2 entry, we use the system as an oracle to generate structured semantic representations. The result of this process is the construction of a new dataset, LC-QuAD 3.0 (LCQ3), which extends LCQ2 with frame-based annotations. This enriched representation now available in all LCQ3 entries not only captures the overall meaning of the question through the identified frame, but also reveals the semantic role of each entity or phrase by linking it to a frame element.
Interestingly, frame elements that remain unfilled (i.e., elements not mapped to any text span in the input) often correspond to the unknowns that the SPARQL query is intended to retrieve—typically, the variables bound in the \textsc{SELECT} clause. For example, in \textsc{SELECT}-type queries, the unmapped frame element is likely to represent the answer to the given question. In contrast, for \textsc{ASK}-type queries, all relevant frame elements tend to be explicitly mentioned in the input, and the goal is to verify the truth value of a fully grounded statement. 

\subsubsection{Stage 4: SPARQL Query Generation}
\begin{figure}
\centering
\includegraphics[width=0.9\textwidth]{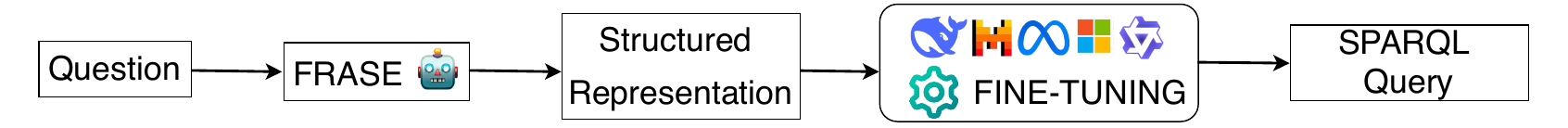}
\caption{Overview of the process of the SPARQL query generation from a question.}
\label{fig:pipeline_simplified}
\end{figure}

All large language models (LLMs) are fine-tuned using an \textit{Instruction-Input-Output} format. In this setup, the instruction indicates to the model that it needs to generate a SPARQL query corresponding to the input question and provides the required context to successfully carry it out. The input consists of the natural language question from LCQ2, optionally augmented with its corresponding structured semantic representation. The output is the SPARQL query that answers the question as show in Figure \ref{fig:pipeline_simplified}.
For experiments involving the structured representation, the frame-based semantic information is appended to the end of the question in the input sequence. This allows the model to condition its generation on both the original question and its enriched semantic context. 

\section{Results and Discussion}
Since LCQ2 does not include annotations for frames or frame-element to argument mappings, the evaluation of our method is performed indirectly through its impact on SPARQL query generation performance. For this evaluation, we report BLEU scores by comparing the generated SPARQL queries with the gold-standard references. Additionally, we use execution-based metrics, namely Accuracy and F1 score, which are computed by executing both the predicted and reference queries against the knowledge base and comparing their returned answers.
For model selection, we focus on several recent and competitive LLMs: LLaMA 3.1 8B, LLaMA 3.2 3B \citep{dubey2024llama}, Phi-4 14B \cite{abdin2024phi4technicalreport}, and DeepSeek-R1 37B \cite{guo2025deepseek}. We also include Mistral 7B GritLM \cite{muennighoff2024generative}, a model with a unique training objective that combines cross-entropy loss with contrastive loss, aiming to optimize both text generation and embedding quality—making it particularly suitable for this test on generalization capability.

To analyze the performance of these models across the different metrics, we evaluate the impact of structured semantic representations on model generalization under two levels of difficulty. The first level is the \textbf{Unknown Template Split} which assesses generalization when the questions structures or patterns encountered at inference time were not present during training—a partial compositional generalization setting. The second is the \textbf{Reformulated Questions}, which represents a more challenging case in which the test questions are naturally phrased and don't follow any defined template.

\subsection{SPARQL Query Generation - Baselines}
Table~\ref{tab:recent-llms-performances} reports the performance of several recent LLMs on the task of SPARQL query generation using the \textbf{Raw Questions} version of the LCQ2 dataset for training. These LLMs constitute our baseline models. To evaluate generalization capabilities, we consider two test scenarios:
(1) the \textbf{Raw Questions} test set, corresponding to template-based questions (left block of Table~\ref{tab:recent-llms-performances}), and (2) the \textbf{Reformulated Questions} test set, composed of human-written paraphrases (right block).
Across all models, we observe that the BLEU scores remain relatively consistent, indicating similar surface-level output quality. However, execution-based metrics such as accuracy and F1 have more variation. Notably, all models exhibit a performance drop when evaluated on the reformulated test set, suggesting limited generalization to linguistic variations not seen during training.
Somewhat unexpectedly, Mistral GritLM \cite{muennighoff2024generative} does not outperform the other models, despite its training objective that combines cross-entropy loss with a contrastive loss component designed to improve both generation and embedding quality. 
Among all models evaluated, Phi-4 achieves the best overall performance, particularly in execution-based metrics. Consequently, we select it for subsequent experiments that analyze the impact of structured frame-based representations on generalization. Given the relatively close performance across models in this task, we hypothesize that trends observed using Phi-4 are likely to extend to other models with comparable capacity.

\begin{table*}
\centering
\begin{threeparttable}
\begin{adjustbox}{width=\linewidth}
\begin{tabular}{cccccccc}
\toprule
\multicolumn{2}{c}{\multirow{2}{*}{\textbf{Models}}} 
& \multicolumn{3}{c}{\textbf{Template-based Questions}} 
& \multicolumn{3}{c}{\textbf{Reformulated Questions}} \\
\cmidrule(lr){3-5}\cmidrule(lr){6-8}
\multicolumn{2}{c}{} 
& \textbf{BLEU} & \textbf{Accuracy} & \textbf{F1} 
& \textbf{BLEU} & \textbf{Accuracy} & \textbf{F1} \\
\midrule
\multirow{6}{*}{LLMs}
& \multirow{1}{*}{Llama 3.1 8B}            & 84 & 30 & 40       & 36 & 13 & 25 \\  
\cmidrule(lr){2-8}
& \multirow{1}{*}{Llama 3.2 3B}            & 83 & 23 & 32       & 37 & 13 & 26 \\  
\cmidrule(lr){2-8}
& \multirow{1}{*}{\textcolor{blue}{Phi-4 14B *}}               & \textcolor{blue}{86} & \textcolor{blue}{37} & \textcolor{blue}{46}       & \textcolor{blue}{42} & \textcolor{blue}{21} & \textcolor{blue}{27} \\  
\cmidrule(lr){2-8}
& \multirow{1}{*}{DeepSeek-R1 37B}         & 86 & 30 & 40       & 41 & 13 & 26 \\ 
\cmidrule(lr){2-8}
& \multirow{1}{*}{Mistral 7B GritLM }         & 85 & 20 & 25       & 43 & 15 & 24 \\
\bottomrule
\end{tabular}
\end{adjustbox}
\begin{tablenotes}
     \item[] \textcolor{blue}{*} The blue-colored line indicates the best overall result and serves as the baseline against which all other configurations will be compared.
  \end{tablenotes}
\end{threeparttable}
\label{tab:recent-llms-performances}
\caption{LLMs comparison in terms of performance (based on BLEU, Accuracy and F1).}
\end{table*}

\subsection{Impact of the structured representations}
As described in Section~\ref{sec:methodology-data-lcq2}, the \textbf{Unknown Template Split} is constructed so that the test set contains only questions whose templates are not seen during training. 
In this experiment, both training and test data are drawn from the \textbf{Unknown Template Split}. For each, we consider two variants: (1) the original natural language questions alone, and (2) the same questions enriched with frame-based semantic annotations. Table~\ref{tab:results-unknown-template-split} reports the results, from which we extract the following key observations:

\paragraph{Raw Questions (without structured augmentation).} As shown in Table ~\ref{tab:results-unknown-template-split}, when models are trained and evaluated using only the \textbf{Raw Questions} on the \textbf{Unknown Template Split}, we observe a substantial drop in BLEU score compared to the \textbf{Original Split} (Table~\ref{tab:recent-llms-performances}). However, execution-based metrics such as accuracy and F1 score slightly improve by approximately 4\%. This counterintuitive result suggests that, although the surface form of generated queries diverges more from the gold standard, the semantic intent is still preserved in many cases. This may be attributed to discrepancies in query lengths across the training, validation, and test sets in the two splits, as illustrated in Table~\ref{tab:lcquad2-splits-stats} and more in detail in Figures ~\ref{fig:splits_stats1} and \ref{fig:splits_stats2} in Appendix \ref{sec:appendix-splits-stats}. In the \textbf{Original Split}, the average query length remains relatively consistent across the training, validation, and test sets. In contrast, in the \textbf{Unknown Template Split}, the average query length in the test set is significantly higher than that of the training and validation sets . As a result, the model tends to generate shorter queries at inference time, which leads to a penalty in BLEU score. However, these shorter queries often contain fewer errors, making them more likely to execute successfully and return more correct answers—thereby improving execution-based metrics such as accuracy and F1.

\paragraph{Frame-Augmented Questions.} When the same experiment is repeated using questions augmented with frame-based structured representations, we observe a notably different behavior. The BLEU score still drops but only by 5 points, indicating improved robustness in surface-level generation. More significantly, execution-based metrics show a marked improvement: accuracy increases by 15\% and F1 score by 19\%. This demonstrates that the structured semantic information introduced by the frames helps the model better capture the intent of the question, even when the surface structure is unfamiliar.


\begin{table*}
    \centering
    
    \begin{threeparttable}
    \begin{adjustbox}{width=\linewidth}
    \begin{tabular}{l|cccccc}
        \toprule
        \diagbox{\textbf{Testing data ↓}}{\textbf{Training data →}} & \multicolumn{6}{c}{\textbf{Raw Questions with Unknown Template Split (UTS)}} \\
        \midrule
        \midrule
        & \multicolumn{3}{c}{-} & \multicolumn{3}{c}{with frames} \\
        \cmidrule(lr){2-4} \cmidrule(lr){5-7}
        & BLEU & Accuracy & F1 & BLEU & Accuracy & F1 \\
        \midrule
        Raw Questions             & 73 & 41 & 50         & -  &  - &  - \\
        Raw Questions with frames & -  & -  & -          & 81 & 52 & 65 \\
        \midrule
        Gain obtained with the use of frames & \multicolumn{6}{c}{BLEU: +8 ---- Accuracy: +11 ---- F1: +15} \\
        \bottomrule
    \end{tabular}
    \end{adjustbox}
    \end{threeparttable}
    \label{tab:results-unknown-template-split}
    \caption{Phi-4 performance on the Unknown Template Split (LCQ2/LCQ3) for different configurations of training.}
\end{table*}

\subsection{Generalization to Template-Free Questions (Reformulated Questions)}
We conducted a comprehensive set of experiments using the three questions variants described in Section~\ref{sec:methodology-data-lcq2}: \textbf{Raw Questions}, \textbf{Reformulated Questions}, and \textbf{Combined Questions} (where both raw questions and  their associated reformulated questions are treated as distinct training examples and mapped to the same SPARQL query). For each dataset variant, we considered both standard and frame-augmented versions, resulting in a total of twelve training configurations.

At inference time, we focus exclusively on the \textbf{Reformulated Questions} version of the test set, using both the original and frame-augmented forms to assess generalization. Tables~\ref{tab:phi4-bleu-performance-lcq2-and-lcq3-reformulated-questions} and~\ref{tab:phi4-acc-f1-performance-lcq2-and-lcq3-reformulated-questions} present the results of these experiments using the Phi-4 model. The key findings are as follows:

\paragraph{Impact of Frame-Based Representations.} Across all dataset variants, incorporating frame-based structured representations consistently improves performance. This trend is clearly visible when comparing the diagonal results within each experimental block—i.e., when both training and testing involve frame augmentation. Models trained and tested with frame-enhanced inputs outperform those using raw questions alone or using frames only at one stage (training or inference), demonstrating the importance of having structured semantic context available throughout the pipeline.

\paragraph{Impact of Combined Questions.}  The best BLEU score is achieved when using the \textbf{Combined Questions} of the LCQ3 dataset—where both template-based and reformulated questions are included and augmented with frame information—at both training and testing time. This setting reflects realistic usage scenarios, where some user queries follow predictable patterns while others are more complex and less structured. While data augmentation from combining question types contributes to improved performance, the addition of frame-based representations has a greater impact, as evidenced by the increase in BLEU score from 67 (combined questions without frames) to 73 (with frames).
A similar trend is observed for accuracy and F1 scores computed from the execution of the generated SPARQL queries. The best results are again obtained with the \textbf{Combined Questions + Frames} configuration, reaching 38\% accuracy and 50\% F1, a substantial improvement compared to the same dataset without frame augmentation (30\% accuracy, 40\% F1). 

Overall, these findings confirm that structured frame-based semantic representations significantly enhance model robustness and generalization to naturally phrased, template-free questions.

\begin{table*}
    \centering
    \begin{threeparttable}
    \begin{adjustbox}{width=\linewidth}
    \begin{tabular}{l|cc|cc|cc}
        \toprule
        \diagbox{\textbf{Testing ↓}}{\textbf{Training →}} & \multicolumn{2}{c|}{\textbf{Raw Questions}} & \multicolumn{2}{c|}{\textbf{Reformulated Questions}} & \multicolumn{2}{c}{\textbf{Combined Questions}} \\
        \midrule
        \midrule
        & - & with frames & - & with frames & - & with frames \\
        \midrule
        Ref Questions               & 42   & 40   & 65    & 53   & 67    & 51 \\
        Ref Questions with frames   & 41   & 54   & 67    & 70   & 60    & \textbf{73} \\
        \midrule
        Gain obtained with frames   & -1   & +14  & +2    & +17  & -7    & +22    \\
        \bottomrule
    \end{tabular}
    \end{adjustbox}
    \end{threeparttable}
    \label{tab:phi4-bleu-performance-lcq2-and-lcq3-reformulated-questions}
    \caption{BLEU performance on reformulated questions for different training configurations.}
\end{table*}

\begin{table*}
    \centering
    \begin{threeparttable}
    \begin{adjustbox}{width=\linewidth}
    \begin{tabular}{l|cccc|cccc|cccc}
        \toprule
        \diagbox{\textbf{Testing ↓}}{\textbf{Training →}} & \multicolumn{4}{c|}{\textbf{Raw Questions}} & \multicolumn{4}{c|}{\textbf{Reformulated Questions}} & \multicolumn{4}{c}{\textbf{Combined Questions}} \\
        \midrule
        \midrule
        & \multicolumn{2}{c}{-} & \multicolumn{2}{c|}{with frames} & \multicolumn{2}{c}{-} & \multicolumn{2}{c|}{with frames} & \multicolumn{2}{c}{-} & \multicolumn{2}{c}{with frames} \\
        \cmidrule(lr){2-3} \cmidrule(lr){4-5} \cmidrule(lr){6-7} \cmidrule(lr){8-9} \cmidrule(lr){10-11} \cmidrule(lr){12-13}   
        & Acc & F1 & Acc & F1 & Acc & F1 & Acc & F1 & Acc & F1 & Acc & F1 \\
        \midrule
        Reformulated Questions             & 14   & 27   & \textcolor{gray}{11}   & \textcolor{gray}{20}   & 20   & 32   & \textcolor{gray}{13}   & \textcolor{gray}{27}   & 30   & 40   & \textcolor{gray}{15}   & \textcolor{gray}{29}   \\
        Reformulated Questions + frames & \textcolor{gray}{13}   & \textcolor{gray}{25}   & 17   & 30   & \textcolor{gray}{17}   & \textcolor{gray}{29}   & 26   & 39   & \textcolor{gray}{21}   & \textcolor{gray}{37}   & \textbf{38}   & \textbf{50}   \\
        \midrule
        Gain obtained with frames     & -1  & -2 & +6 & +10 & -3 & -3 & +13 & +12 & -9 & -3 & +23 & +21   \\
        \bottomrule
    \end{tabular}
    \end{adjustbox}
    \end{threeparttable}
    \label{tab:phi4-acc-f1-performance-lcq2-and-lcq3-reformulated-questions}
    \caption{Accuracy and F1 performance with reformulated questions for different training configurations.}
\end{table*}

\section{Conclusion}
\label{sec:conclusion}
In this paper, we introduced \textbf{FRASE}, a novel approach designed to provide structured semantic representations of natural language questions in order to improve the generalization capabilities of large language models (LLMs) in the task of SPARQL query generation. \textbf{FRASE} addresses the brittleness of LLMs to lexical variation and syntactic reformulation—two key challenges in question-to-query tasks—by leveraging frame semantics as an intermediate representation.
Extensive experiments conducted across multiple recent LLMs show that FRASE consistently improves SPARQL generation performance, especially in generalization settings involving unseen templates and naturally reformulated questions. These findings demonstrate the potential of integrating structured semantic knowledge into LLM-driven systems to enhance their robustness and abstraction capabilities.
This work opens a promising avenue for research at the intersection of semantic parsing and prompt engineering. While we focused on question answering, the broader insight is that structured semantic enhancement can benefit a wide range of natural language tasks. In future work, we aim to explore the application of \textbf{FRASE} beyond question-to-query translation, treating prompts as general cases of natural language inputs that could similarly benefit from semantic structuring.

\section{Limitations}
While our proposed method improves the performance compared to baseline models on Wikidata, we acknowledge some limitations. First, the best obtained F1 score remain low (50 F1-score) and a further analysis should explore which cases are not well-handled by our frame-based representations. 
Second, although the approach is designed to be KB-agnostic in principle, we have not yet evaluated it on knowledge bases beyond Wikidata. Its effectiveness in other settings remains to be validated.
Third, the success of our frame-based alignment depends heavily on the availability and quality of textual descriptions for relations and classes within the KB. In cases where such descriptions are missing, sparse, or poorly written, the semantic search used for frame detection may yield suboptimal or noisy alignments. This limitation is particularly relevant for incomplete or less curated knowledge bases, where relation descriptions may be inconsistent or unavailable.
Finally, our current implementation relies exclusively on the English version of FrameNet, which limits the applicability of the method to English-language questions and KBs. Extending the approach to multilingual settings would require either high-quality multilingual FrameNet resources or robust cross-lingual mapping strategies, which are non-trivial and beyond the scope of this study.

\section*{Acknowledgments}
We are grateful to the NSERC Discovery Grant Program, which has funded this research. The authors would also like to express their gratitude to Compute Canada (Calcul Quebec) for providing computational resources.

\bibliography{colm2025_conference}
\bibliographystyle{colm2025_conference}

\appendix
\section{Splits Statistics} \label{sec:appendix-splits-stats}
\begin{figure}
\centering
\includegraphics[width=1\textwidth]{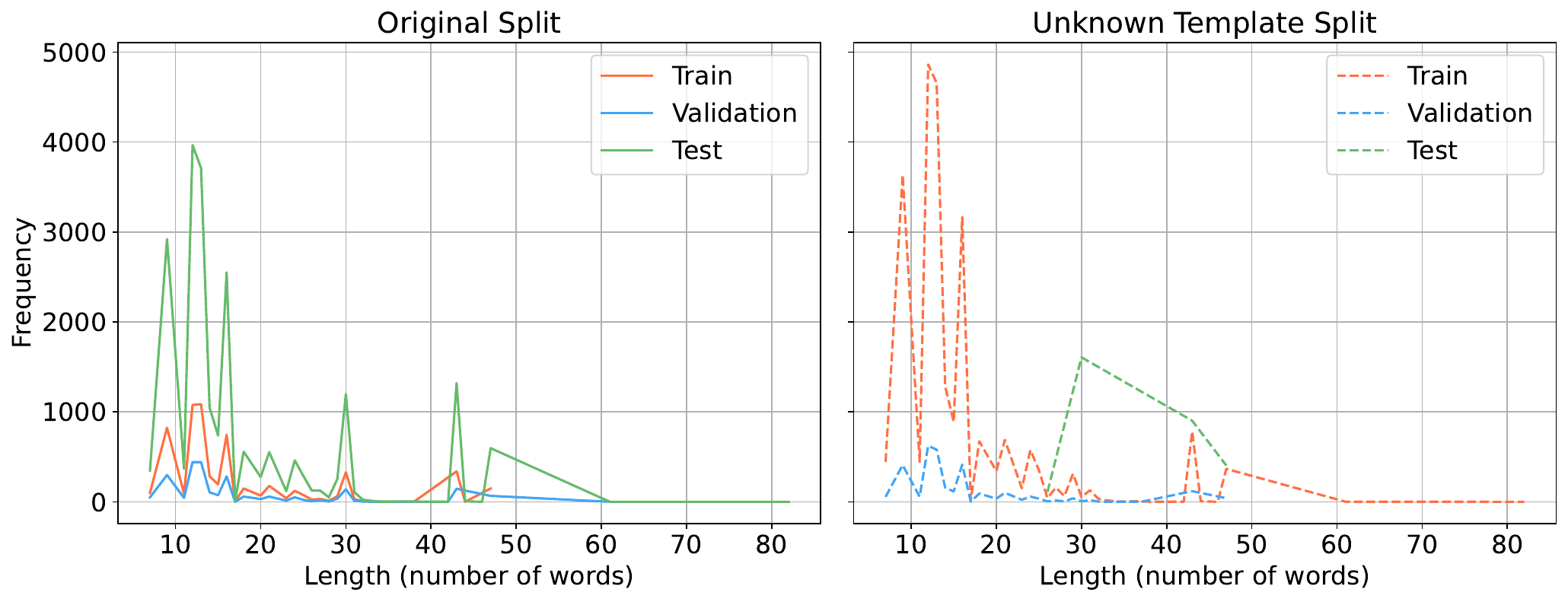}
\caption{Distribution of query length (number of words) in Original Split and Unknown Template Split.}
\label{fig:splits_stats1}
\end{figure}

\begin{figure}
\centering
\includegraphics[width=1\textwidth]{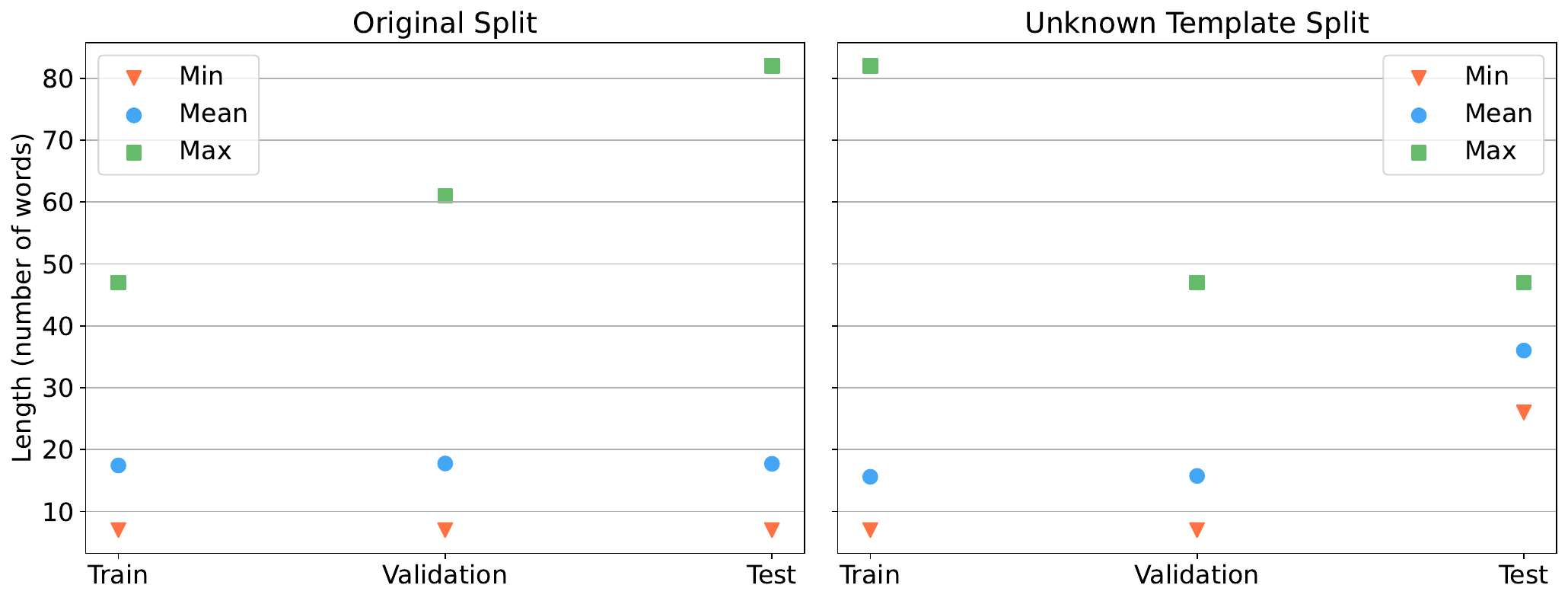}
\caption{AVG/Min/Max query length (number of words) in Original Split and Unknown Template Split.}
\label{fig:splits_stats2}
\end{figure}

Figures~\ref{fig:splits_stats1} and~\ref{fig:splits_stats2} show the distribution of question lengths (in number of words) and the corresponding statistics (minimum, mean, maximum) for the training, validation, and test sets in the two LC-QuAD 2.0 splits: the \textbf{Original Split} and the \textbf{Unknown Template Split}. In the \textbf{Original Split}, the length distributions are relatively consistent across all subsets, with similar averages, indicating structural alignment between training and test data. In contrast, the \textbf{Unknown Template Split} reveals a clear discrepancy: test questions are, on average, significantly longer than those in the training and validation sets. This shift reflects the intended challenge of the split, where models must generalize to unseen question templates, which tend to be more complex and verbose. Such distributional differences likely contribute to the performance drop observed in this setting, particularly for surface-level generation metrics like BLEU.

\section{FRASE additional detail}
The Algorithm \ref{algo:frase} shows the Stage 1 of \textbf{FRASE} that detects the frames evoked in any LCQ2 question as depicted in first part of Figure \ref{fig:example_lcq2_question_annotated_by_frase}.

\begin{algorithm}\label{algo:frase}
\caption{Identify Frames for LCQ2 Questions}
\label{alg:identify_frames}
\begin{algorithmic}[1]
\Require LCQ2\_question (natural language question), SPARQL\_query (corresponding SPARQL query), VectorDatabase (stores frame vectors), KB (Knowledge Base with ontology and descriptions)
\Ensure EvokedFrames (set of frames evoked by the question)
\State Initialize an empty set \texttt{EvokedFrames}
\State \textbf{Preprocess Frames:}
\For{each frame in the set of available frames}
    \State Represent the frame using its name, description, and list of frame-elements
    \State Encode the frame representation into a vector using the embedding-model
    \State Store the frame vector in \texttt{VectorDatabase}
\EndFor
\State \textbf{Extract Relevant KB Elements:}
\State Parse \texttt{SPARQL\_query} to extract relevant KB element identifiers (URIs) corresponding to relation or class
\State \textbf{Generate KB Element Representations:}
\For{each URI in the extracted URIs}
    \State Fetch the label and textual description of the corresponding KB element from \texttt{KB}
    \State Encode the label and description using the embedding-model
\EndFor
\State \textbf{Align KB Elements with Frames:}
\For{each vector representation of a KB element}
    \State Perform a similarity search in \texttt{VectorDatabase} to find the most similar frame vector(s)
    \If{a match is found (similarity score $\geq$ threshold)}
        \State Add the matched top-\(k=1\)frame(s) to \texttt{EvokedFrames}
    \EndIf
\EndFor
\State \Return \texttt{EvokedFrames}
\end{algorithmic}
\end{algorithm}

\section{LCQ2 Questions Annotation by \textbf{FRASE}}\label{sec:annotation_lcq2_by_frase}
\begin{figure}
\centering
\includegraphics[width=1\textwidth]{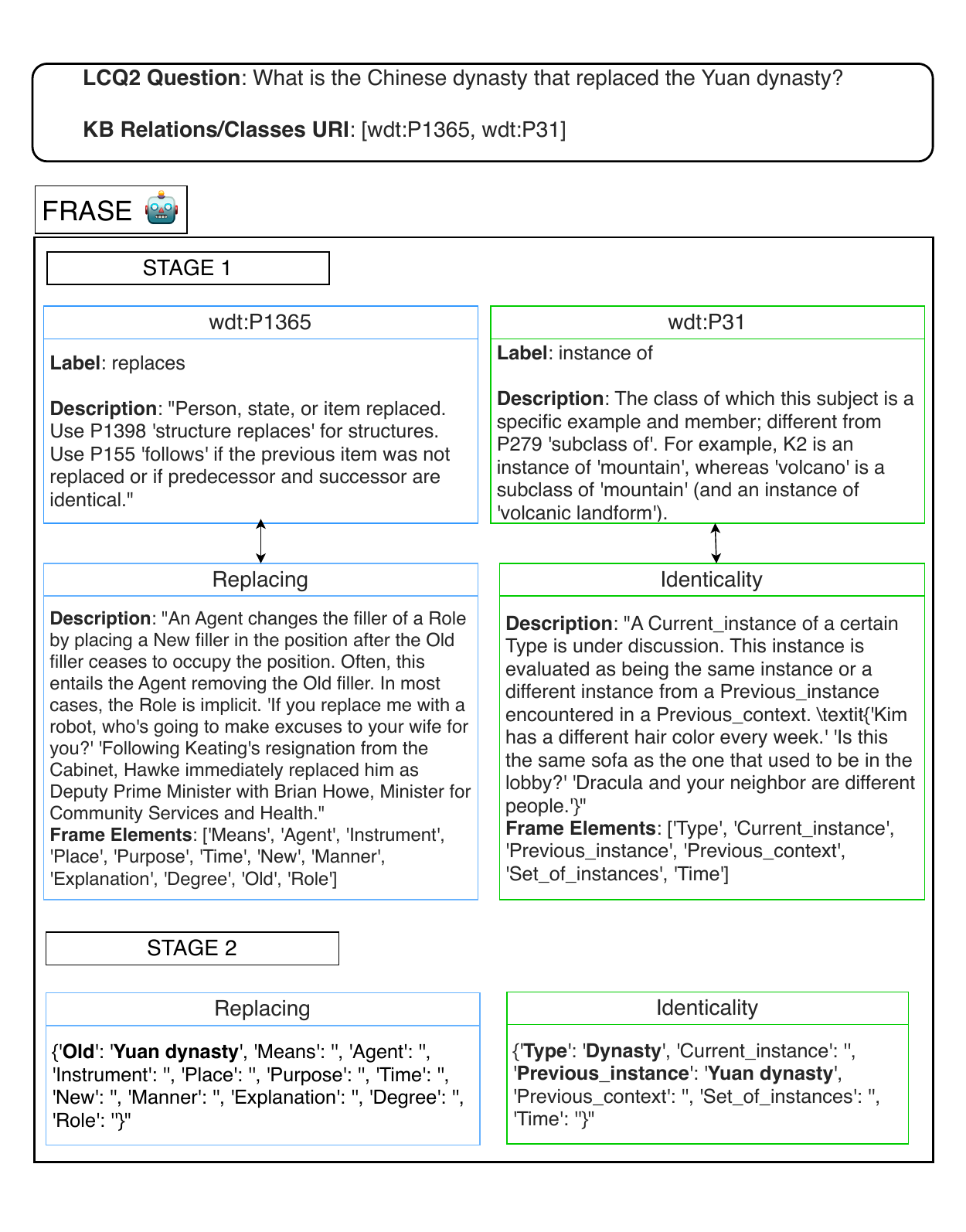}
\caption{An example of frame-based structured representation of text semantic.}
\label{fig:example_lcq2_question_annotated_by_frase}
\end{figure}

Figure~\ref{fig:example_lcq2_question_annotated_by_frase} illustrates how our proposed \textbf{FRASE} pipeline semantically enriches a natural language question from LC-QuAD 2.0 using frame-based structured representations. In \textbf{Stage 1}, each relation URI in the associated SPARQL query is aligned with a corresponding FrameNet frame based on textual similarity. For instance, the relation 'wdt:P1365' ("replaces") is aligned with the \textbf{Replacing} frame, and "wdt:P31" ("instance of") is mapped to \textbf{Identicality}. In \textbf{Stage 2}, the system identifies the relevant \textbf{Frame Elements} and links them to corresponding spans in the question text. In this example, the element "Old" is mapped to "Yuan dynasty", and "Type" is inferred as "Dynasty". This structured representation captures the underlying semantic roles involved in the question and provides an interpretable abstraction that can be used to improve SPARQL generation and generalization.

\section{Experimental details}
We fine-tune all models using the QLoRA method, which combines 4-bit quantization with parameter-efficient fine-tuning. Specifically, we quantize the base model weights using NF4 quantization and bfloat16 computation via the \texttt{BitsAndBytesConfig}. We then apply a low-rank adaptation (LoRA) on key components of the transformer layers (e.g., \texttt{q\_proj}, \texttt{k\_proj}, \texttt{v\_proj}, etc.) with rank 16, \texttt{lora\_alpha} 16, and no dropout. The fine-tuning is performed using the \texttt{adamw\_8bit} optimizer with a learning rate of \(2 \times 10^{-4}\), and linear scheduling. We save and evaluate the model at the end of each epoch and report the best checkpoint based on validation loss. Further training parameters are detailed in Table~\ref{tab:fine-tuning-technical-details}.

\begin{table*}
    \centering
    \begin{adjustbox}{width=\linewidth}
    \begin{threeparttable}
    \begin{tabular}{lcc}
        \toprule
        \textbf{Parameters} & \textbf{Values} \\
        \midrule
        Max Sequence Length & 2048 \\
        Packing & False (for faster training) \\
        Per Device Batch Size & 8 \\
        Gradient Accumulation Steps & 4 \\
        Warmup Steps & 5 \\
        Number of Epochs & 10 (adjustable) \\
        Learning Rate & 2e-4 \\
        Precision Mode & bfloat16 \\
        Quantization Type & 4-bit (NF4) \\
        LoRA Rank ($r$) & 16 \\
        LoRA Alpha & 16 \\
        LoRA Dropout & 0 \\
        Target Modules & Attention + MLP Projections\tnote{1} \\
        Optimizer & adamw\_8bit \\
        Weight Decay & 0.01 \\
        Learning Rate Scheduler & Linear \\
        Random Seed & 1618 \\
        Evaluation Strategy & Epoch \\
        \bottomrule
    \end{tabular}
    \begin{tablenotes}
        \item[1] \texttt{q\_proj}, \texttt{k\_proj}, \texttt{v\_proj}, \texttt{o\_proj}, \texttt{gate\_proj}, \texttt{up\_proj}, \texttt{down\_proj}
    \end{tablenotes}
    \end{threeparttable}
    \end{adjustbox}
    \label{tab:fine-tuning-technical-details}
    \caption{Technical details of the fine-tuning}
\end{table*}

\end{document}